\pdfoutput=1

\documentclass[11pt]{article}

\usepackage{acl}
\usepackage{times}
\usepackage{latexsym}
\usepackage{amsmath}

\usepackage[T1]{fontenc}

\usepackage[utf8]{inputenc}

\usepackage{microtype}

\usepackage{inconsolata}

\usepackage{graphicx}
\usepackage{caption} 

\title{LT4SG@SMM4H’24: Tweets Classification for Digital Epidemiology of Childhood Health Outcomes Using Pre-Trained Language Models}

\author{
  \parbox[t]{0.3\textwidth}{%
    \centering
    Dasun Athukoralage \\
    \textnormal{NirvanaClouds} \\
    \textnormal{dasun@nirvanaclouds.com}
  }
  \hspace{0.1\textwidth} 
  \parbox[t]{0.3\textwidth}{%
    \centering
    Thushari Atapattu \\
    \textnormal{University of Adelaide} \\
    \textnormal{thushari.atapattu@adelaide.edu.au}
  }
  \\\\
  \parbox[t]{0.32\textwidth}{%
    \centering
    \textbf{Menasha Thilakaratne} \\
    \textnormal{University of Adelaide} \\
    \textnormal{menasha.thilakaratne@adelaide.edu.au}
  }
  \hspace{0.1\textwidth} 
  \parbox[t]{0.3\textwidth}{%
    \centering
    \textbf{Katrina Falkner} \\
    \textnormal{University of Adelaide} \\
    \textnormal{katrina.falkner@adelaide.edu.au}
    }
}

\begin{document}
\maketitle
\begin{abstract}
This paper presents our approaches for the SMM4H’24 Shared Task 5 on the binary classification of English tweets reporting children’s medical disorders. Our first approach involves fine-tuning a single RoBERTa-large model, while the second approach entails ensembling the results of three fine-tuned BERTweet-large models. We demonstrate that although both approaches exhibit identical performance on validation data, the BERTweet-large ensemble excels on test data. Our best-performing system achieves an F1-score of 0.938 on test data, outperforming the benchmark classifier by 1.18\%. 
\end{abstract}

\section{Introduction \& Motivation}

Chronic childhood disorders like attention-deficit/hyperactivity disorder (ADHD), autism spectrum disorders (ASD), delayed speech, and asthma significantly impact a child's development and well-being, often extending into adulthood. Approximately 1 in 6 (17\%) children aged 3-17 years in the United States experience a developmental disability, with ADHD, ASD, and others contributing to this statistic \citep{zab2019preva}. In previous studies \citep{guntuku2019language, hswen2019using, edo2019twitter}, Twitter data have been utilized to identify self-reports of the aforementioned disorders; however, the identification of reports concerning these disorders in users’ children has not been explored. It may be of interest to explore Twitter's potential in continuing to collect users' tweets postpartum, enabling the detection of outcomes in childhood.

\section{Task and Data Description}

\subsection{Task}

The SMM4H-2024 workshop and shared tasks have a special focus on Large Language Models (LLMs) and generalizability for natural language processing (NLP) in social media. We participated in Task 5, which is \textquotesingle Binary classification of English tweets reporting children’s medical disorders\textquotesingle\/. The objective is to automatically differentiate tweets from users who have disclosed their pregnancy on Twitter and mention having a child with ADHD, ASD, delayed speech, or asthma (annotated as "1"), from tweets that merely refer to a disorder (annotated as "0").

\subsection{Data}

There were three different datasets provided: training, validation, and test datasets. The training and validation datasets were labeled while the test dataset was not. All datasets are composed entirely of tweets posted by users who had reported their pregnancy on Twitter, that report having a child with a disorder and tweets that merely mention a disorder. The training, validation, and test sets contain 7398 tweets, 389 tweets, and 1947 tweets, respectively.

\section{Methodology}
\subsection{Baseline}
A benchmark classifier, based on a RoBERTa-large model \citep{liu2019roberta}, has achieved an F1-score of 0.927 for the \textquotesingle positive\textquotesingle\ class (i.e., tweets that report having a child with a disorder) on the test data for Task 5 \citep{klein2024using}.

\subsection{Models Used}
We investigated three Transformer based models which are BioLinkBERT-large \citep{yasunaga2022linkbert}, RoBERTa-large and BERTweet-large \citep{nguyen2020a}. BioLinkBERT was selected for its specialized understanding of biomedical NLP tasks, RoBERTa for its domain-independent NLP capabilities, and BERTweet for its superior performance in Tweet-specific NLP tasks. We fine-tuned each model with the training dataset and evaluated its performance using the validation dataset.

\subsection{Training Regime}
The complete description of the training regime is in Appendix~\ref{appendix:A}. Primary hyperparameters including learning rate, weight decay, and batch size were determined as described in the subsequent section.

\subsection{Hyperparameter Optimization}
We conducted hyperparameter optimization that relied on HuggingFace’s Trainer API with the Ray Tune backend \citep{liaw2018tune}. Details about the hyperparameters are described in Appendix~\ref{appendix:B}.

\section{Preliminary Experiments}

Each selected model was trained for 3 iterations, with 10 epochs per iteration. At the end of each epoch, its F1-score was recorded. The F1-score for each model was determined based on its performance with the validation dataset. We saved the best-performing epoch (i.e., the best F1-score for the positive class) for each model in each iteration. The results are shown in Appendix~\ref{appendix:C}.

As shown in \hyperref[appendix:C]{Table 4}, RoBERTa-large and BERTweet-large perform similarly on the validation dataset, and considerably better than BioLinkBERT-large, even though it has been pre-trained on a large corpus of biomedical data. Therefore, we decided to remove BioLinkBERT-large to carry out further experiments for this task \citep{guo2020benchmarking}.

\subsection{Ensembling Strategy}
The issue arises when fine-tuning large-transformer models on small datasets: the classification performance varies significantly with slightly different training data and random seed values, even when using the same hyperparameter values \citep{dodge2020fine}. To overcome this high variance and provide more robust predictions, we propose ensembles of multiple fine-tuned RoBERTa-large models and BERTweet-large models separately. We create two separate ensemble models using the best models corresponding to three iterations for each RoBERTa-large and BERTweet-large. All three iterations use the same hyperparameters, and only differ in the initial random seed. A hard majority voting mechanism combines the predictions of these models (see Appendix~\ref{appendix:D}).

\begin{table}[htbp] 
  \centering
  \resizebox{\columnwidth}{!}{%
  \begin{tabular}{llll}
    \hline
    \textbf{Classifier} & \textbf{F1-score} & \textbf{Precision} & \textbf{Recall} \\
    \hline
    RoBERTa-large Ensemble & 0.934783 & 0.914894 & \textbf{0.955556} \\
    BERTweet-large Ensemble & \textbf{0.945055} & \textbf{0.934783} & \textbf{0.955556} \\

    \hline
  \end{tabular}
  }
  \captionsetup{labelformat=empty} 
  \caption{Table 1: Performance results for ensemble classifiers on validation data.}
\end{table}

As shown in Table 1, the BERTweet-large ensemble performs better than the RoBERTa-large ensemble. This is fundamentally because its performance variation is less for three iterations, as indicated in \hyperref[appendix:C]{Table 4}. Another noteworthy observation is that the performance of the BERTweet-large ensemble is identical to that of the best iteration (\hyperref[appendix:C]{Table 4}, 2\textsuperscript{nd} run) of RoBERTa-large (see Appendix~\ref{appendix:E}). \hyperref[appendix:F]{Figure 1} shows the corresponding confusion matrices for both classifiers which are also identical.

\section{Results and Conclusion}
Since RoBERTa-large best-run and BERTweet-large ensemble are performing equally well on the validation data, we tested the performance of both classifiers on unseen, unlabeled test data. As shown in Table 2, the BERTweet-large ensemble classifier outperforms the mean and median performance on the test data among all teams’ submissions by a considerable margin, as well as the benchmark classifier by 1.18\%. Additionally, we can observe that even though both classifiers perform equally well on validation data, the BERTweet-large ensemble model performs significantly better on test data. One possible reason for this is that different runs of BERTweet-large might excel at capturing different aspects of the data or learning different patterns.

\begin{table}[htbp] 
  \centering
  \resizebox{\columnwidth}{!}{%
  \begin{tabular}{llll}
    \hline
    \textbf{Model} & \textbf{F1-score} & \textbf{Precision} & \textbf{Recall} \\
    \hline
    Baseline & 0.927 & 0.923 & 0.940 \\
    Mean & 0.822 & 0.818 & 0.838 \\
    Median & 0.901 & 0.885 & 0.917 \\
    RoBERTa-large best-run & 0.925 & 0.908 & 0.942 \\
    \textbf{BERTweet-large Ensemble} & \textbf{0.938} & \textbf{0.930} & \textbf{0.946} \\

    \hline
  \end{tabular}
  }
  \captionsetup{labelformat=empty} 
  \caption{Table 2: Results for our two proposed approaches on the test data, including the mean, median, and baseline scores.}
\end{table}

When fine-tuning complex pre-trained language models, one issue on small datasets is the instability of the classification performance. To overcome this, we combined the predictions of multiple BERTweet-large models in an ensemble. By doing so, we achieved significantly better results in terms of F1-score for SMM4H'24 Task 5. For future work, it's interesting to investigate how the system performance varies when adding more BERTweet-large iterations (i.e., runs) to the ensemble.

\bibliography{custom}

\clearpage

\appendix

\section{Training Regime}
\label{appendix:A}

Experiments were conducted using Google Colab Pro+ equipped with an NVIDIA A100 Tensor Core GPU boasting 40 gigabytes of available GPU RAM. The Hugging Face Transformers Python library \citep{wolf2019transformers} and its Trainer API facilitated training procedures. Each model was trained on the training datasets for 3 iterations and 10 epochs per iteration. We used HuggingFace's Trainer Class's default \textquotesingle AdamW\textquotesingle\ and \textquotesingle linear warmup with cosine decay\textquotesingle\ as the optimizer and scheduler respectively. The maximum sequence length for all models was set to 512. FP-16 mixed precision training was employed to enable larger batch sizes and expedited training.

\section{Hyperparameters}
\label{appendix:B}

We utilized Ray Tune's built-in "BasicVariantGenerator" algorithm\footnote{\url{https://docs.ray.io/en/latest/tune/api/doc/ray.tune.search.basic_variant.BasicVariantGenerator.html}} for hyperparameter search, paired with the First-In-First-Out (FIFO) scheduler. Since BasicVariantGenerator has the ability to dynamically generate hyperparameter configurations based on predefined search algorithms (e.g., random search, Bayesian optimization), it enables more efficient exploration of the search space. The hyperparameters optimized using BasicVariantGenerator are presented in Table 3.

\begin{table}[htbp] 
  \centering
  \resizebox{\columnwidth}{!}{%
  \begin{tabular}{llll}
    \hline
    \textbf{Model} & \textbf{Learning Rate} & \textbf{Weight Decay} & \textbf{Batch Size} \\
    \hline
    BioLinkBERT-large & 6.10552e-06 & 0.00762736 & 16 \\
    RoBERTa-large & 7.21422e-06 & 0.00694763 & 8 \\
    BERTweet-large & 1.17754e-05 & 0.01976150 & 8 \\

    \hline
  \end{tabular}
  }
  \captionsetup{labelformat=empty} 
  \caption{Table 3: Hyperparameters optimized via BasicVariantGenerator.}
\end{table}

\section{The F1 Scores of Selected Models}
\label{appendix:C}

\begin{table}[htbp] 

  \centering
  \resizebox{\columnwidth}{!}{%
  \begin{tabular}{llllll}
    \hline
    \textbf{Model} & \textbf{1\textsuperscript{st} run
} & \textbf{2\textsuperscript{nd} run
} & \textbf{3\textsuperscript{rd} run
} & \textbf{Mean F1} & \textbf{SD} \\
    \hline
    BioLinkBERT-large & 0.855019 & \textbf{0.875969} & 0.863159 & 0.864716 & 0.010561 \\
    RoBERTa-large & 0.931408 & \textbf{0.945055} & 0.931408 & 0.935957 & 0.007879 \\
    BERTweet-large & \textbf{0.940741} & 0.934307 & 0.933824 & \textbf{0.936291} & \textbf{0.003862} \\

    \hline
  \end{tabular}
  }
  \captionsetup{labelformat=empty} 
  \caption{Table 4: The F1 scores of the BioLinkBERT-large, RoBERTa-large, and BERTweet-large classifiers on the validation data. The mean F1 score and standard deviation are also provided.}
  
\end{table}

\vspace{\baselineskip} 
\vspace{\baselineskip} 
\vspace{\baselineskip} 
\vspace{\baselineskip} 

\section{A Hard Majority Voting Mechanism}
\label{appendix:D}

A hard majority voting mechanism combines the predictions of three fine-tuned BERTweet-large models:

\begin{equation}
\hat{y} = \underset{c}{\arg\max} \sum_{i=1}^{n} \mathbf{1}(\hat{y}_i = c)
\end{equation}

where $\mathbf{1}(\cdot)$ represents the indicator function, which returns either \textquotesingle 1\textquotesingle\ or \textquotesingle 0\textquotesingle\ for the class label $c$ predicted by the $i$-th classifier.

\vspace{\baselineskip} 

\section{RoBERTa-large best-run vs the BERTweet-large Ensemble}
\label{appendix:E}

\begin{table}[htbp] 
  \centering
  \resizebox{\columnwidth}{!}{%
  \begin{tabular}{llll}
    \hline
    \textbf{Classifier} & \textbf{F1-score} & \textbf{Precision} & \textbf{Recall} \\
    \hline
    RoBERTa-large best-run & \textbf{0.945055} & \textbf{0.934783} & \textbf{0.955556} \\
    BERTweet-large Ensemble & \textbf{0.945055} & \textbf{0.934783} & \textbf{0.955556} \\

    \hline
  \end{tabular}
  }
  \captionsetup{labelformat=empty} 
  \caption{Table 5: Performance comparison of the RoBERTa- large best-run vs the BERTweet-large ensemble on vali- dation data.}
\end{table}

\section{Confusion Matrices}
\label{appendix:F}

\begin{figure}[htbp]
  \centering
  \includegraphics[width=\columnwidth]{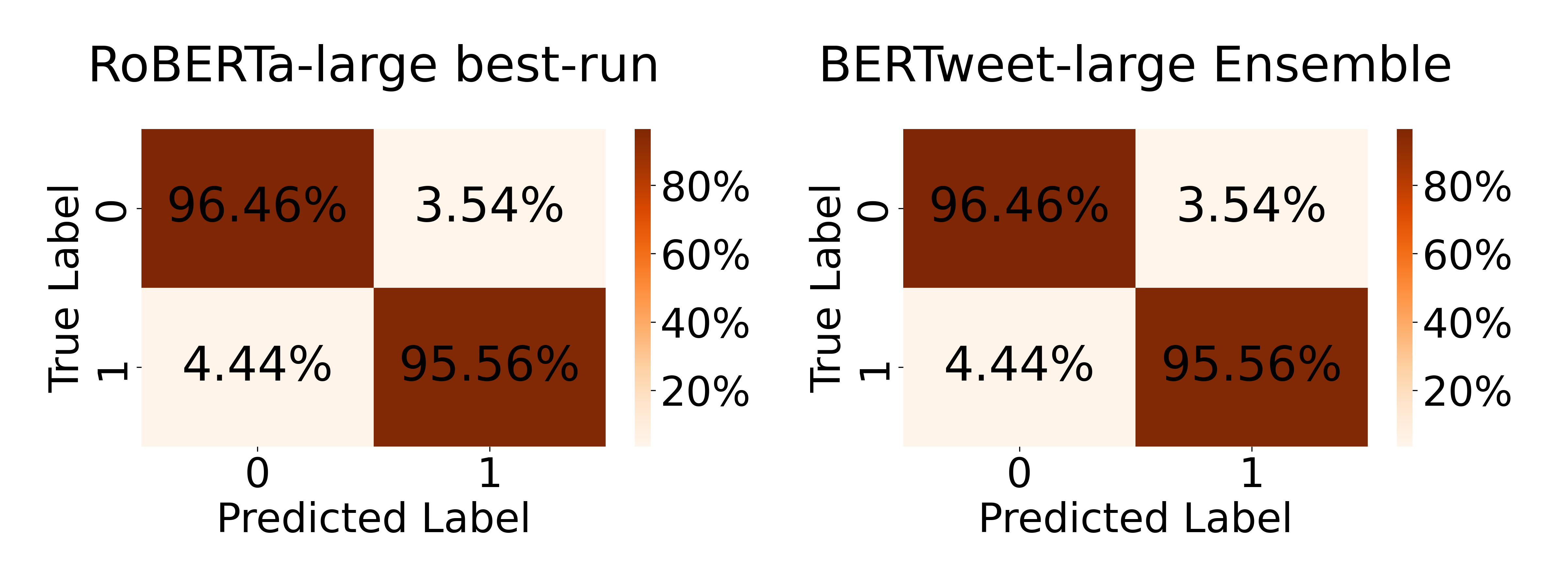} 
  \caption{Confusion matrices of the RoBERTa-large best-run and BERTweet-large ensemble on the validation dataset.}
  \label{fig:example}
\end{figure}

\end{document}